%% file: main.tex
\documentclass[10pt,twocolumn]{article}

\usepackage[a4paper,margin=0.72in,columnsep=0.24in]{geometry}
\usepackage{booktabs}
\usepackage{graphicx}
\usepackage{amsmath,amssymb}
\usepackage{microtype}
\usepackage{url}
\usepackage[hidelinks]{hyperref}
\usepackage{xcolor}
\usepackage{tikz}
\usetikzlibrary{arrows.meta,positioning,fit}

\newcommand{\method}{ShiftSplit-AD}
\newcommand{\R}{\mathbf{R}}
\newcommand{\Lmat}{\mathbf{L}}
\newcommand{\Smat}{\mathbf{S}}
\newcommand{\Emat}{\mathbf{E}}

\title{ShiftSplit-AD: Separating Domain Shift from Defects in Foundation-Feature Visual Anomaly Detection}
\author{
MUHAMATHU AMEER ALI AACAAS MUHAMATH\\
Department of Electrical Engineering\\
University of Moratuwa\\
Moratuwa, Sri Lanka\\
\texttt{muhamathmaaa.23@uom.lk}
}
\date{}

\begin{document}
\maketitle

\begin{abstract}
Visual anomaly detectors based on frozen foundation-model features commonly score distances from test patches to a memory of normal features. Benign acquisition changes can also enlarge these distances, confounding domain variation with defects. We investigate whether structured decomposition of nearest-normal DINOv2 residuals can suppress shift-induced evidence while retaining unseen defects. \method{} decomposes the patch residual matrix into low-rank and row-sparse components and scores the sparse component, with an optional low-rank/sparse fusion. The experiments expose a central trade-off rather than a universal separation: genuine defects can contain correlated, low-dimensional structure, so filtering broad residual activity may also remove defect information. On AeBAD-S, using settings fixed after Bottle development, sparse-only scoring improves image AUROC from 0.6780 to 0.7294 and AUPRC from 0.8052 to 0.8465. Paired bootstrap 95\% intervals for the improvements are [0.0238, 0.0808] and [0.0170, 0.0650], respectively. However, sparse-only scoring reduces mean clean AUROC from 0.9890 to 0.9133 on four held-out MVTec categories and degrades Bottle localization. These findings show that residual decomposition can help when domain shift strongly contaminates anomaly evidence, but preserving defect structure remains the limiting problem.
\end{abstract}

\input{sections/01_introduction}
\input{sections/02_related_work}
\input{sections/03_method}
\input{sections/04_experiments}
\input{sections/05_results}
\input{sections/06_discussion}
\input{sections/07_limitations}
\input{sections/08_conclusion}

\bibliographystyle{plain}
\bibliography{references}
\end{document}

%% file: sections/01_introduction.tex
\section{Introduction}

Industrial visual anomaly detection is commonly posed as a one-class problem: normal images are available for training, whereas the appearance of future defects is unknown. Feature-memory methods address this setting by extracting pretrained patch representations and measuring the distance from each test patch to a bank of normal features. Such systems are simple and effective on established benchmarks such as MVTec AD~\cite{bergmann2019mvtec}, and memory-based designs including SPADE~\cite{cohen2020spade} and PatchCore~\cite{roth2022patchcore} demonstrate the strength of local pretrained features.

Distance from normality, however, does not identify the cause of that distance. Illumination, viewpoint, background, blur, and sensor noise can move a normal image away from the training distribution. A nearest-normal detector may therefore produce a large residual for a harmless acquisition change. This ambiguity matters in deployment because the operating domain need not reproduce the conditions under which the normal memory was collected. AeBAD was introduced specifically to expose this issue for aero-engine blade inspection, with changes in illumination, view, alignment, and scale between training and testing~\cite{zhang2023aebad}.

We study this problem directly in frozen foundation-feature space. DINOv2 provides general visual features without task-specific fine-tuning~\cite{oquab2024dinov2}. For each test patch, we subtract its nearest normal-memory feature and stack the residuals into a matrix. The central question is whether the structure of that matrix can help distinguish broad domain variation from localized defect evidence.

We study \method{}, a lightweight post-processing step that approximates the residual matrix by low-rank and row-sparse components. Low-rank/sparse decomposition itself is classical~\cite{candes2011rpca}, and anomaly detection under distribution shift is an established research problem~\cite{cao2023adshift,carvalho2023invariant,zhang2023aebad}. The specific contribution examined here is the use of \emph{per-test-image nearest-normal foundation-feature residual structure} to study shift suppression versus defect preservation. Sparsity acts on spatial rows rather than individual entries. The initial intuition is that correlated, spatially broad residual activity may be absorbed into the low-rank term, while concentrated activity remains in the row-sparse term.

The experiments do not support the simplistic identity ``domain shift equals low rank, defects equal sparse.'' Real defects also contain correlated structure, and removing it can lower detection and localization quality. Instead, the results reveal a trade-off between suppressing shift-induced residual activity and preserving defect information. Sparse-only scoring strongly improves Bottle under synthetic Gaussian noise and improves AeBAD-S overall, but it loses accuracy on held-out MVTec categories whose raw residual detector is already strong. A fusion of low-rank and sparse scores recovers much of this loss without becoming universally superior.

Our contributions are threefold. First, we instantiate and empirically study per-test-image low-rank/row-sparse decomposition of frozen DINO nearest-normal patch residuals, rather than claiming novelty for decomposition or shift-aware anomaly detection in general. Second, we evaluate the method using a transparent development/held-out protocol: Bottle informs the fixed decomposition, top-5 aggregation, and fusion weight; four other MVTec categories and AeBAD-S are subsequently evaluated without retuning these settings. Third, we report positive and negative evidence together, including paired bootstrap uncertainty on AeBAD-S, a defect-type failure on breakdown, and substantial localization degradation. The resulting claim is deliberately conditional: structured residual filtering can help when domain variation contaminates raw anomaly evidence, but its usefulness depends on whether removed structure is nuisance or defect signal.

%% file: sections/02_related_work.tex
\section{Related Work}

\paragraph{Industrial visual anomaly detection.}
MVTec AD established a diverse benchmark with normal-only training data, anomalous test images, and pixel-level masks~\cite{bergmann2019mvtec}. Many high-performing methods build on pretrained representations. SPADE retrieves similar normal images and uses deep pyramid correspondences for localization~\cite{cohen2020spade}. PaDiM models a multivariate Gaussian distribution of pretrained features at each spatial location~\cite{defard2020padim}. PatchCore stores locally aware normal patch features and reduces memory through coreset selection~\cite{roth2022patchcore}. Our raw baseline belongs to this broad memory-based family but uses frozen DINOv2 patch features and exact nearest-neighbor residuals. The repository also contains a PatchCore-style DINO coreset comparison; because it changes the backbone and does not reproduce all canonical PatchCore components, we do not label it as canonical PatchCore.

\paragraph{Foundation features for anomaly detection.}
DINOv2 learns general-purpose visual representations through large-scale self-supervised training~\cite{oquab2024dinov2}. Its patch tokens provide a spatial representation suitable for correspondence and anomaly maps. Vision-language models provide another route to general anomaly recognition; for example, WinCLIP adapts CLIP to zero- and few-shot anomaly classification and segmentation through prompt ensembles and window features~\cite{jeong2023winclip}. \method{} is narrower: it does not use language or learn a task-specific model. It tests whether the residual geometry of one frozen visual foundation model can be reweighted after nearest-normal retrieval.

\paragraph{Domain shift.}
Standard anomaly benchmarks emphasize category and defect diversity, while deployment may also change the distribution of normal data. ADShift formalizes anomaly detection under distribution shift and reduces gaps between in-distribution and shifted normal features during training and inference~\cite{cao2023adshift}. Invariant Anomaly Detection uses causal reasoning to derive a regularizer that learns partial distribution invariance across environments~\cite{carvalho2023invariant}. AeBAD explicitly introduces real acquisition shift for aero-engine blades and attributes major changes to view and illumination, together with scale and alignment variation~\cite{zhang2023aebad}.

\method{} differs from these approaches in intervention point and supervision. ADShift performs distribution alignment, whereas \method{} leaves the frozen representation and normal memory unchanged and decomposes each test image's nearest-normal residual matrix. Invariant Anomaly Detection learns partially invariant representations through regularization, whereas \method{} is post-hoc and training-free after memory construction. We use AeBAD-S as the principal real-shift evaluation rather than treating synthetic corruptions as a substitute. Controlled brightness, contrast, blur, and noise remain diagnostics of false-positive evidence, not models of the full acquisition domain.

\paragraph{Subspace and residual methods.}
Recent training-free work makes the relationship between foundation features and subspaces especially relevant. SubspaceAD fits a PCA model to patch features from a few normal DINOv2 images and scores reconstruction residuals from that learned \emph{normal} subspace~\cite{lendering2026subspacead}. By contrast, \method{} first performs nearest-neighbor matching to a potentially large normal memory and then decomposes the $256\times384$ residual matrix of each individual test image; its low-rank component is test-conditioned rather than a PCA model of normal training features.

SPARC uses up to eight verified-normal images from the incoming target lot to estimate spatially indexed nuisance subspaces and projects them out before a frozen detector~\cite{han2026sparc}. \method{} assumes no target-normal calibration images: it estimates structure independently within each test residual matrix. SPARC is therefore a few-shot deployment-calibration method, while \method{} studies target-free, query-specific decomposition.

Anomaly-Related Residual Fields studies residual evolution in diffusion models and combines a source-calibrated extractor with cross-domain field alignment~\cite{gao2026residualfields}. Its residuals, dynamics, and transfer setting differ from the static DINO nearest-neighbor patch residuals used here. \method{} neither follows diffusion trajectories nor aligns source and target fields; it asks whether one test image's frozen-feature residual geometry supports useful reweighting.

\paragraph{Low-rank and sparse decomposition.}
Principal Component Pursuit decomposes a matrix into low-rank and entry-sparse terms under assumptions that enable exact recovery~\cite{candes2011rpca}. Our use is heuristic and does not inherit those guarantees. Residual rows correspond to spatial patches, so we employ an $\ell_{2,1}$ row-group penalty rather than entrywise $\ell_1$ sparsity. More importantly, neither component has a known semantic identity. The decomposition is used to probe and reweight residual structure, not to assert a generative law equating low rank with domain shift.

%% file: sections/03_method.tex
\section{Method}

\subsection{Frozen Patch Features and Normal Memory}

Let a frozen DINOv2-small encoder map an image to $P=256$ patch embeddings $\{\mathbf{z}_i\}_{i=1}^{P}$, arranged on a $16\times16$ grid. Each embedding has dimension $D=384$. All patch embeddings from normal training images form a memory bank $\mathcal{M}\subset\mathbb{R}^{D}$. No anomalous test image enters this memory.

For a test patch, exact Euclidean nearest-neighbor matching gives
\begin{equation}
 \mathbf{m}_i^*=\arg\min_{\mathbf{m}\in\mathcal{M}}
 \lVert\mathbf{z}_i-\mathbf{m}\rVert_2.
\end{equation}
The residual and raw patch anomaly score are
\begin{equation}
 \mathbf{r}_i=\mathbf{z}_i-\mathbf{m}_i^*,\qquad
 a_i^R=\lVert\mathbf{r}_i\rVert_2.
\end{equation}
Stacking rows yields $\R=[\mathbf{r}_1^\top;\ldots;\mathbf{r}_P^\top]\in\mathbb{R}^{P\times D}$.

\subsection{ShiftSplit Decomposition}

Using a classical low-rank/sparse regularization pattern~\cite{candes2011rpca}, we seek an approximate per-test-image decomposition $\R\approx\Lmat+\Smat$ through
\begin{equation}
\min_{\Lmat,\Smat}\ \frac{1}{2}\lVert\R-\Lmat-\Smat\rVert_F^2
+\lambda_L\lVert\Lmat\rVert_*
+\lambda_S\sum_{i=1}^{P}\lVert\Smat_i\rVert_2.
\label{eq:objective}
\end{equation}
The nuclear norm encourages correlated residual structure to concentrate in a low-dimensional subspace. The row-group penalty encourages only a subset of spatial patches to remain active in $\Smat$.

Starting from zeros, the implementation alternates for a fixed number of iterations:
\begin{align}
 \Lmat &\leftarrow \operatorname{SVT}_{\lambda_L}(\R-\Smat),\\
 \Smat_i &\leftarrow \left(1-\frac{\lambda_S}{\lVert(\R-\Lmat)_i\rVert_2}\right)_+
 (\R-\Lmat)_i.
\end{align}
$\operatorname{SVT}$ soft-thresholds singular values. The second update is group soft thresholding. The remaining reconstruction component is $\Emat=\R-\Lmat-\Smat$. We use $\lambda_L=10$, $\lambda_S=8$, and 20 iterations throughout the locked evaluations.

\subsection{Scoring and Fusion}

Sparse-only patch scores are $a_i^S=\lVert\Smat_i\rVert_2$. We also study a patch-level fusion
\begin{equation}
 a_i(\alpha)=\alpha\lVert\Smat_i\rVert_2+(1-\alpha)\lVert\Lmat_i\rVert_2.
\end{equation}
The image score for all final experiments is the mean of the five largest patch scores. Bottle experiments motivated top-5 aggregation and selected $\alpha=0.75$ as a development choice. This weight is fixed for the four held-out MVTec categories and AeBAD-S; it is not claimed to be universally optimal.

\begin{figure*}[t]
\centering
\resizebox{\textwidth}{!}{%
\begin{tikzpicture}[
  node distance=5mm and 5mm,
  box/.style={draw,rounded corners,align=center,minimum height=8mm,minimum width=19mm,fill=blue!5,font=\small},
  split/.style={draw,rounded corners,align=center,minimum height=8mm,minimum width=18mm,fill=orange!10,font=\small},
  arrow/.style={-{Latex[length=2mm]},thick}
]
\node[box] (image) {Input\\image};
\node[box,right=of image] (dino) {Frozen\\DINOv2};
\node[box,right=of dino] (features) {$16\!\times\!16$ patch\\features $\mathbf{z}_i$};
\node[box,right=of features] (nn) {Nearest normal\\memory match};
\node[box,right=of nn] (residual) {Residual matrix\\$\R\in\mathbb{R}^{256\times384}$};
\node[split,right=of residual,yshift=6mm] (low) {Low-rank $\Lmat$};
\node[split,right=of residual,yshift=-6mm] (sparse) {Row-sparse $\Smat$};
\node[box,right=11mm of sparse] (patch) {Sparse or fused\\patch scores};
\node[box,right=of patch] (topk) {Top-5\\mean};
\node[box,right=of topk] (score) {Image anomaly\\score};
\draw[arrow] (image)--(dino); \draw[arrow] (dino)--(features); \draw[arrow] (features)--(nn); \draw[arrow] (nn)--(residual);
\draw[arrow] (residual)--(low); \draw[arrow] (residual)--(sparse);
\draw[arrow] (low.east)--++(5mm,0)|-(patch.north west); \draw[arrow] (sparse)--(patch);
\draw[arrow] (patch)--(topk); \draw[arrow] (topk)--(score);
\end{tikzpicture}
}
\caption{\method{} pipeline. Frozen patch features are matched to a normal-only memory. The resulting residual matrix is decomposed and reweighted before top-5 image aggregation. The decomposition components are structural variables, not guaranteed semantic labels for shift and defect.}
\label{fig:method}
\end{figure*}
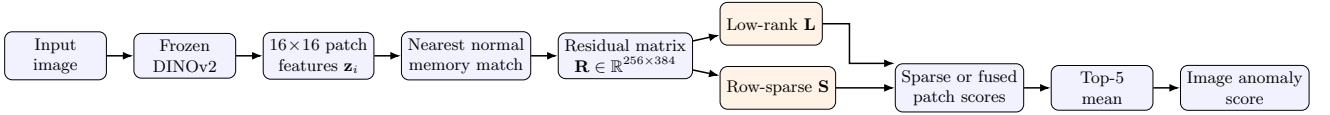

%% file: sections/04_experiments.tex
\section{Experimental Setup}

\subsection{Development and Evaluation Protocol}

Protocol transparency is essential because several design choices arose during exploration. MVTec Bottle is the development and diagnostic category. The thresholds $\lambda_L=10$ and $\lambda_S=8$ arose from a Bottle residual study; top-5 image aggregation was motivated by a Bottle scoring ablation; and fusion weight $\alpha=0.75$ was selected on Bottle. Bottle test labels therefore informed development, and Bottle results are not presented as untouched generalization.

After these choices, Cable, Hazelnut, Metal Nut, and Wood are treated as held-out MVTec categories for the fusion experiment. AeBAD-S is subsequently evaluated with the same decomposition, aggregation, and fusion settings and without AeBAD-specific retuning. This sequence supports transfer-oriented evidence, but it is not equivalent to a preregistered benchmark evaluation.

\subsection{Datasets}

MVTec AD~\cite{bergmann2019mvtec} provides normal-only training images and test images with pixel masks for 15 object and texture categories. The completed multi-category experiments use Bottle, Cable, Hazelnut, Metal Nut, and Wood. For the controlled shift protocol, only normal test images are transformed while defects remain clean, isolating false-positive pressure. The symmetric protocol transforms both classes.

AeBAD-S~\cite{zhang2023aebad} is the real-domain-shift benchmark. The executed protocol uses 521 normal training images to build a $133{,}376\times384$ patch memory. The test set contains 1,639 images: 490 normal and 1,149 anomalous, divided into ablation (169), breakdown (329), fracture (389), and groove (262). These counts are verified from the saved per-image prediction file.

\subsection{Shifts, Baselines, and Metrics}

Synthetic diagnostics apply brightness and contrast factors, Gaussian blur, and pixelwise Gaussian noise. The main Bottle comparison uses noise standard deviation 20. A severity sweep and eight noise seeds probe stability. The raw baseline is exact DINOv2-small nearest-normal patch matching. The S-only method uses $\lVert\Smat_i\rVert_2$, and fusion combines $\Lmat$ and $\Smat$ with $\alpha=0.75$. A PatchCore-style DINO coreset baseline is included only as a diagnostic implementation; it should not be interpreted as a canonical reproduction of PatchCore~\cite{roth2022patchcore}.

Image-level performance is measured by AUROC and, for AeBAD-S, AUPRC. Localization uses pixel AUROC and AUPRC after bilinear upsampling from the $16\times16$ patch grid. AeBAD marginal confidence intervals use 2,000 nonparametric bootstrap resamples. Direct comparisons use 5,000 paired bootstrap resamples, with the same resampled image indices applied to both methods. We report the empirical distribution of paired differences and avoid interpreting an empirical zero tail count as a literal probability of zero.

%% file: sections/05_results.tex
\section{Results}

\subsection{Real Domain Shift on AeBAD-S}

\input{tables/aebad_overall}

Table~\ref{tab:aebad-overall} gives the principal real-shift result. S-only \method{} raises AUROC by 0.0514 and AUPRC by 0.0412 over raw residual scoring. Its marginal 95\% bootstrap intervals are [0.7034, 0.7556] for AUROC and [0.8240, 0.8686] for AUPRC, compared with [0.6493, 0.7064] and [0.7792, 0.8328] for raw scoring.

The paired bootstrap is more directly relevant to improvement. For S-only minus raw, the AUROC difference is 0.0514 with a 95\% interval of [0.0238, 0.0808]; the AUPRC difference is 0.0412 with [0.0170, 0.0650]. Every one of 5,000 saved AUROC resamples had a positive difference, while 99.94\% of AUPRC resamples were positive. These are empirical bootstrap frequencies, not proof that the population probability of non-improvement is zero.

\begin{figure}[t]
\centering
\includegraphics[width=\columnwidth]{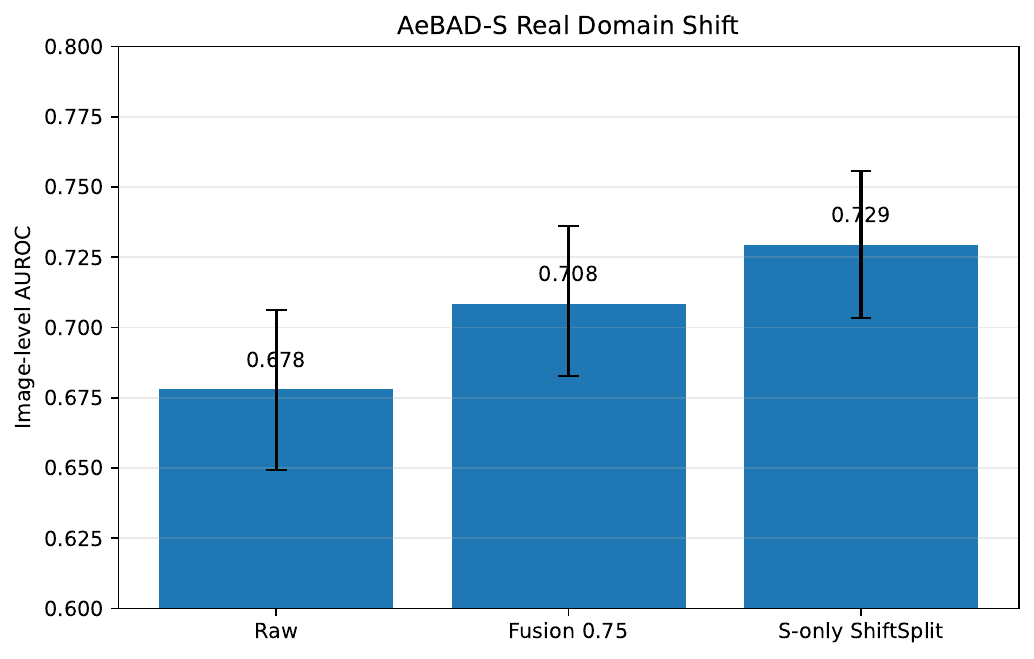}
\caption{AeBAD-S image AUROC with marginal 95\% bootstrap intervals. Both decomposition scores improve over raw residual ranking, with S-only highest.}
\label{fig:aebad-overall}
\end{figure}

\input{tables/aebad_defects}

The aggregate gain is heterogeneous (Table~\ref{tab:aebad-defects}). S-only improves both metrics for ablation, fracture, and groove, but breakdown AUROC falls from 0.8894 to 0.7796 and AUPRC from 0.7931 to 0.6803. Fusion is also below raw on breakdown. Thus, the overall improvement does not justify a universal claim across defect morphologies.

\begin{figure}[t]
\centering
\includegraphics[width=\columnwidth]{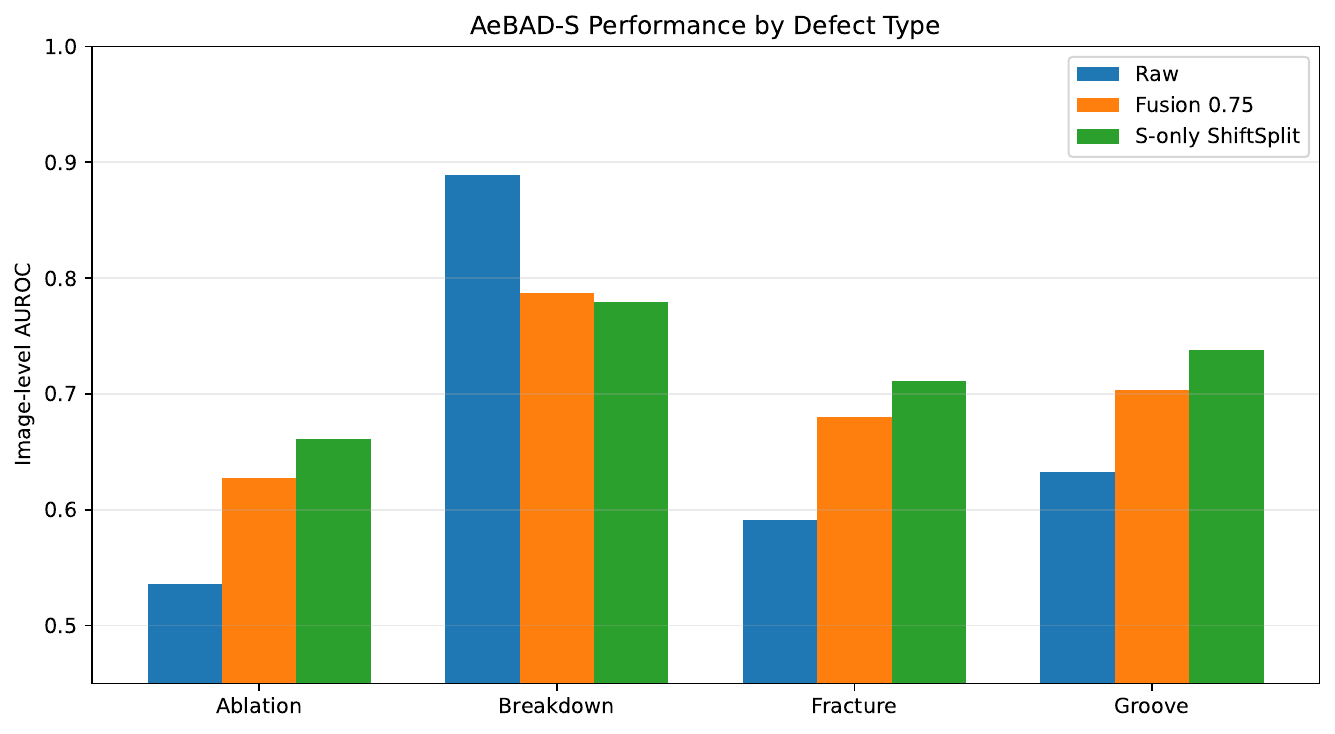}
\caption{AeBAD-S image AUROC by defect type. Breakdown is the central negative case: residual structure useful to the raw detector is suppressed by decomposition.}
\label{fig:aebad-types}
\end{figure}

\subsection{Held-Out MVTec Categories}

\input{tables/mvtec_heldout}

The held-out MVTec results reverse the AeBAD ordering (Table~\ref{tab:heldout}). Raw DINO residuals are already near ceiling under clean and noise-20 conditions. S-only filtering removes useful evidence, reducing the four-category mean by 0.0757 clean and 0.0946 under noise. Fusion recovers much of this loss, reaching 0.9740 and 0.9626, but remains below raw on average. Wood clean is the only category/condition in which fusion exceeds raw (0.9939 versus 0.9833).

\begin{figure*}[t]
\centering
\includegraphics[width=.485\textwidth]{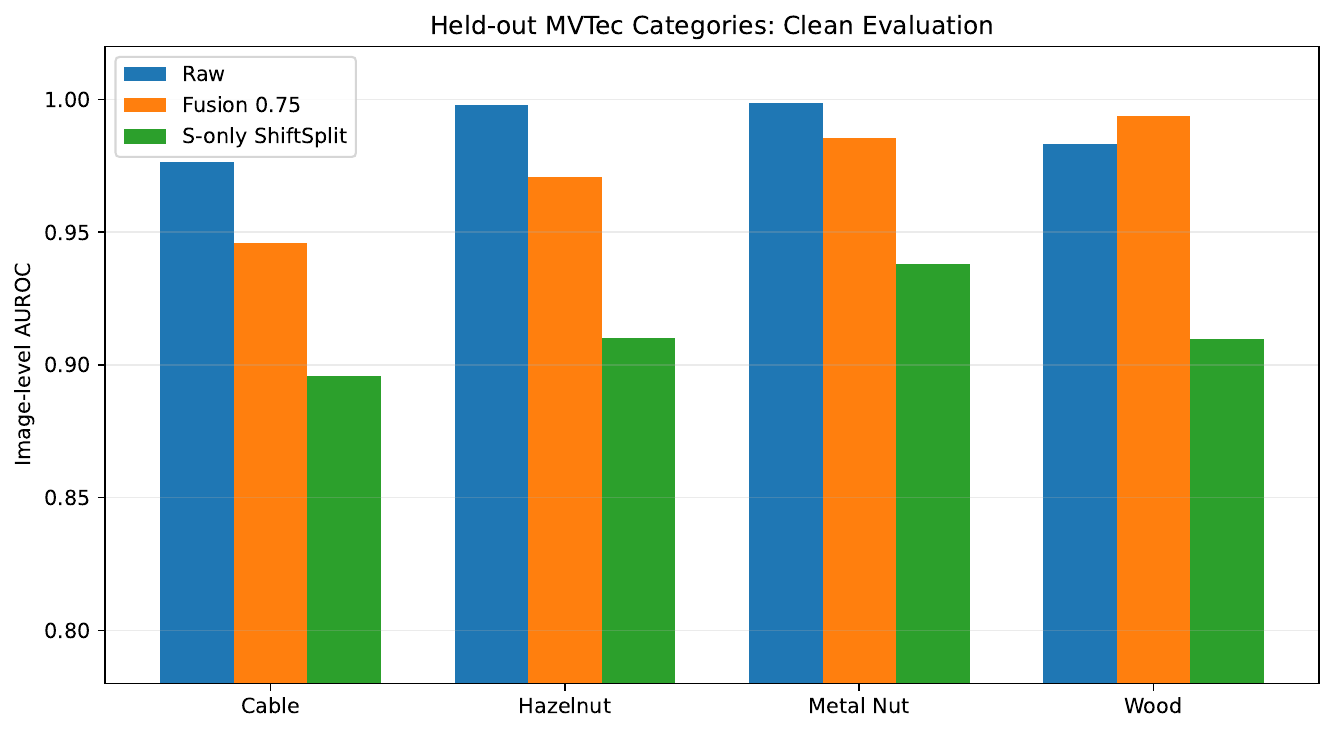}\hfill
\includegraphics[width=.485\textwidth]{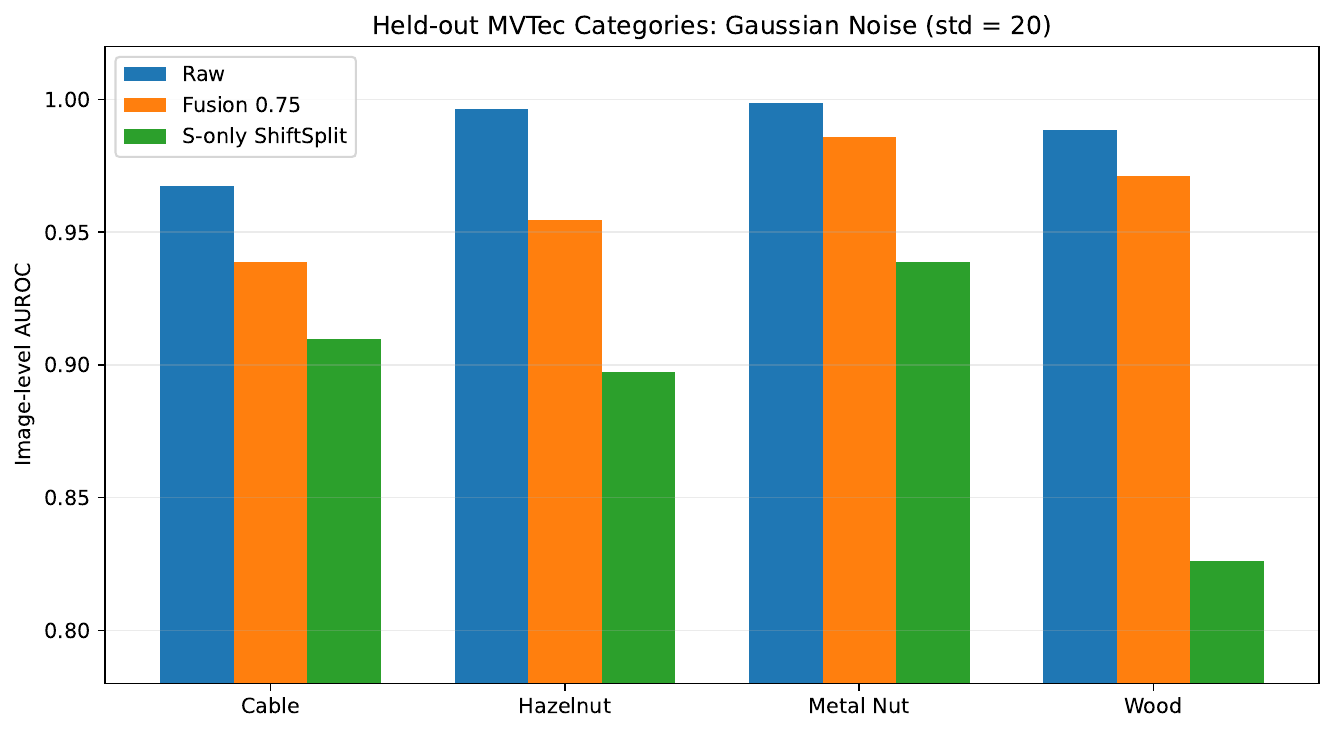}
\caption{Held-out MVTec image AUROC after Bottle selection of $\alpha=0.75$. Left: clean. Right: Gaussian noise with standard deviation 20. Fusion consistently retains more defect evidence than S-only, but raw scoring remains strongest in seven of eight category/condition comparisons.}
\label{fig:heldout}
\end{figure*}

Across all five evaluated MVTec categories, including Bottle, mean clean AUROC is 0.9912 raw versus 0.9272 S-only. Mean noise AUROC is 0.9386 raw versus 0.9118 S-only. Bottle is the exception driving the apparent synthetic-shift benefit: under top-5 noise-20 scoring, it improves from 0.7421 to 0.9865.

\subsection{Bottle Diagnostics and Ablations}

\input{tables/ablations}

Component ablation supports the trade-off interpretation. On Bottle, raw $\R$ and low-rank $\Lmat$ both achieve 1.0000 clean AUROC, while $\Smat$ reaches 0.9825 and the remainder $\Emat$ 0.5325. Under noise-20, $\Smat$ reaches 0.9865, compared with 0.7421 for $\R$, 0.6325 for $\Lmat$, and 0.5079 for $\Emat$. The useful defect signal is therefore not confined to $\Smat$, even though $\Smat$ is most robust in this corruption.

Noise-seed results show that Bottle improvement is not attributable to one draw. Across eight seeds, at standard deviations 10, 20, and 30, raw mean AUROC is 0.5862, 0.6452, and 0.5875, whereas S-only reaches 0.9522, 0.9576, and 0.9472. In the symmetric noise-20 protocol, which shifts both normal and defective images, raw/S-only AUROC is 0.7262/0.9198. Under controlled noise-20, a threshold calibrated to 5\% clean-normal false-positive rate produces a raw false-positive rate of 1.00 and S-only rate of 0.05. These Bottle diagnostics explain why sparse filtering helps: noise broadly elevates raw evidence on normal images.

The same filtering harms localization. The per-image CSV averages over 63 Bottle defects give pixel AUROC 0.9815 raw versus 0.8964 S-only, and pixel AUPRC 0.7357 versus 0.3951. The result is consistent with defect residuals being partly assigned to $\Lmat$.

\begin{figure*}[t]
\centering
\includegraphics[width=.92\textwidth]{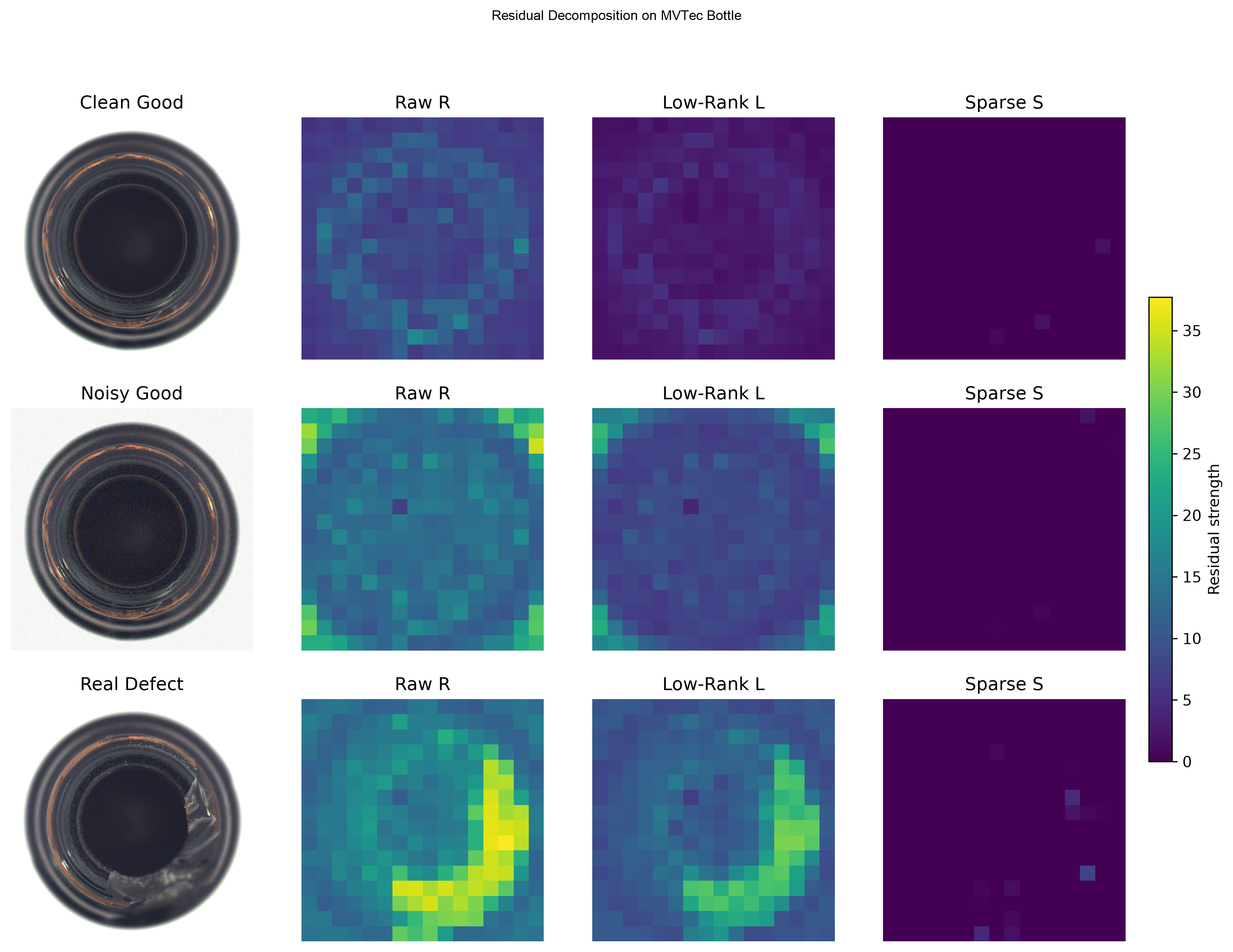}
\caption{Bottle development visualization for a clean normal, noisy normal, and real defect. Noise creates broad raw activity that is largely absorbed outside $\Smat$. The defect also has substantial low-rank structure, illustrating why S-only scoring can lose detection and localization information. This figure is diagnostic rather than evidence of a universal component semantics.}
\label{fig:decomposition}
\end{figure*}

%% file: tables/aebad_overall.tex
\begin{table}[t]
\centering
\caption{AeBAD-S image-level results. Confidence intervals are percentile bootstrap intervals over 2,000 valid resamples.}
\label{tab:aebad-overall}
\small
\begin{tabular}{lcc}
\toprule
Method & AUROC & AUPRC \\
\midrule
Raw residual & 0.6780 & 0.8052 \\
Fusion ($\alpha=0.75$) & 0.7084 & 0.8253 \\
S-only \method{} & \textbf{0.7294} & \textbf{0.8465} \\
\bottomrule
\end{tabular}
\end{table}

%% file: tables/aebad_defects.tex
\begin{table}[t]
\centering
\caption{AeBAD-S defect-type results. Each defect type is ranked against the same 490 normal test images. S-only helps three types but harms breakdown.}
\label{tab:aebad-defects}
\setlength{\tabcolsep}{3.4pt}
\scriptsize
\begin{tabular}{lrrrr}
\toprule
& \multicolumn{2}{c}{AUROC} & \multicolumn{2}{c}{AUPRC} \\
Defect & Raw & S-only & Raw & S-only \\
\midrule
Ablation  & 0.5356 & \textbf{0.6610} & 0.2497 & \textbf{0.3776} \\
Breakdown & \textbf{0.8894} & 0.7796 & \textbf{0.7931} & 0.6803 \\
Fracture  & 0.5914 & \textbf{0.7110} & 0.4695 & \textbf{0.6385} \\
Groove    & 0.6329 & \textbf{0.7379} & 0.4088 & \textbf{0.5838} \\
\bottomrule
\end{tabular}
\end{table}

%% file: tables/mvtec_heldout.tex
\begin{table}[t]
\centering
\caption{Mean image AUROC over four held-out MVTec categories after choosing $\alpha=0.75$ on Bottle. Noise uses Gaussian standard deviation 20.}
\label{tab:heldout}
\small
\begin{tabular}{lcc}
\toprule
Method & Clean & Noise \\
\midrule
Raw residual & \textbf{0.9890} & \textbf{0.9877} \\
S-only \method{} & 0.9133 & 0.8931 \\
Fusion ($\alpha=0.75$) & 0.9740 & 0.9626 \\
\bottomrule
\end{tabular}
\end{table}

%% file: tables/ablations.tex
\begin{table}[t]
\centering
\caption{Bottle development ablations using top-5 scoring. Localization values are means of the 63 per-image metrics saved in the result CSV.}
\label{tab:ablations}
\setlength{\tabcolsep}{3.5pt}
\scriptsize
\begin{tabular}{lcc}
\toprule
Experiment / method & Raw & S-only \\
\midrule
Clean image AUROC & \textbf{1.0000} & 0.9825 \\
Noise-20 image AUROC & 0.7421 & \textbf{0.9865} \\
Pixel AUROC (mean) & \textbf{0.9815} & 0.8964 \\
Pixel AUPRC (mean) & \textbf{0.7357} & 0.3951 \\
\bottomrule
\end{tabular}
\end{table}

%% file: sections/06_discussion.tex
\section{Discussion}

\paragraph{Low rank is not a domain label.}
Dataset diagnostics show that 90\% of residual spectral energy requires on average 67.55 singular components for clean Bottle normals, 48.10 for noisy normals, and 42.92 for real defects. Defects are at least as spectrally concentrated as the synthetic shift under this diagnostic. Consequently, singular-value concentration alone cannot distinguish shift from defect, and the components should not be assigned fixed semantics.

\paragraph{Spatial breadth is useful but conditional.}
The number of patches required for 80\% of residual energy averages 158.5 for clean normals, 173.9 for noisy normals, and 142.4 for defects. Noise is spatially broader in this development setting. Row shrinkage exploits that distinction and explains the large Bottle noise gain. But this is a tendency, not an invariant: large defects can be broad, and local glare or reflection can be spatially sparse.

\paragraph{Suppression versus preservation.}
The method's behavior is best understood as reweighting. S-only scoring suppresses broad correlated evidence aggressively. This helps on AeBAD-S overall and on three defect groups, suggesting that domain-induced residual structure interferes with ranking there. It hurts held-out MVTec and AeBAD breakdown, where the raw representation already separates classes or where the removed component contains useful defect information. Fusion moves toward a compromise: it consistently improves over S-only on held-out MVTec averages, yet retains less shift suppression than S-only on AeBAD-S.

\paragraph{What the bootstrap establishes.}
The paired intervals indicate that the observed AeBAD-S aggregate improvement is stable to resampling this test set. They do not establish universality across future domains, defect mixtures, or alternative development choices. Defect-type decomposition is particularly important because a positive aggregate difference coexists with a large negative difference for breakdown.

\paragraph{Practical implication.}
Residual decomposition may be most useful as a robustness control when normal validation data reveals domain-induced false positives. Deploying S-only scoring by default is not supported. A safer future system could adapt the mixture of raw, low-rank, and sparse evidence using normal calibration data or an explicit estimate of shift severity, without consulting anomalous evaluation labels.

%% file: sections/07_limitations.tex
\section{Limitations}

The study uses one frozen backbone, DINOv2-small, so the findings may depend on its representation and preprocessing. The five MVTec categories do not cover the full benchmark, and Bottle was heavily used during development. The held-out labels were not used to choose $\alpha$, but the overall research process is exploratory rather than preregistered.

The decomposition parameters are fixed and manually motivated. They do not adapt to category, domain, defect scale, or memory quality. The alternating proximal procedure also leaves nonzero reconstruction error, and no theoretical recovery guarantee applies to these feature residuals.

AeBAD-S supplies real domain variation, but it represents one component family and one data collection setting. More real-shift datasets are needed. The AeBAD analysis is image-level; a corresponding real-shift localization study is not reported.

Localization on Bottle degrades substantially under S-only scoring. The $16\times16$ patch grid and bilinear upsampling already limit boundary precision, while the decomposition removes correlated defect structure. Large/global defects, sparse reflections, dense sensor noise, and viewpoint changes may violate the working spatial intuition in different ways.

The normal memory uses exact patch retrieval and can be expensive. Chunked distance computation controls peak memory but not asymptotic work. Global nearest-neighbor patch retrieval does not enforce spatial correspondence, so a test patch may match a normal patch from a different object location; this may increase tolerance to pose variation but can also suppress spatially meaningful anomalies.

The study focuses on mechanism analysis rather than broad state-of-the-art comparison; several recent distribution-shift and subspace-based methods are discussed but not reproduced under the same protocol. Finally, the PatchCore-style DINO coreset baseline in the repository is useful diagnostically but is not a canonical PatchCore reproduction and is therefore not used to make broad comparative claims.

%% file: sections/08_conclusion.tex
\section{Conclusion}

We investigated whether structured decomposition of frozen nearest-normal patch residuals can reduce domain-shift-induced anomaly evidence. \method{} separates a residual matrix into low-rank and row-sparse components and scores the sparse component or a low-rank/sparse fusion. The results reject a simple semantic split between shift and defect. Instead, they expose a trade-off: aggressive sparse scoring can suppress broad nuisance activity, but it can also discard correlated defect information.

With Bottle-derived settings fixed, S-only scoring improves AeBAD-S AUROC from 0.6780 to 0.7294 and AUPRC from 0.8052 to 0.8465, with positive paired bootstrap intervals. At the same time, it underperforms raw scoring on held-out MVTec averages, hurts AeBAD breakdown, and degrades Bottle localization. The appropriate conclusion is therefore conditional rather than universal. Structured residual filtering is promising when domain variation strongly contaminates raw residuals; reliable deployment will require a principled mechanism for balancing shift suppression against defect preservation.

%% file: references.bib
@article{oquab2024dinov2,
  title   = {{DINOv2}: Learning Robust Visual Features without Supervision},
  author  = {Oquab, Maxime and Darcet, Timoth{\'e}e and Moutakanni, Th{\'e}o and Vo, Huy V. and Szafraniec, Marc and Khalidov, Vasil and Fernandez, Pierre and Haziza, Daniel and Massa, Francisco and El-Nouby, Alaaeldin and Assran, Mahmoud and Ballas, Nicolas and Galuba, Wojciech and Howes, Russell and Huang, Po-Yao and Li, Shang-Wen and Misra, Ishan and Rabbat, Michael and Sharma, Vasu and Synnaeve, Gabriel and Xu, Hu and J{\'e}gou, Herv{\'e} and Mairal, Julien and Labatut, Patrick and Joulin, Armand and Bojanowski, Piotr},
  journal = {Transactions on Machine Learning Research},
  year    = {2024}
}

@inproceedings{bergmann2019mvtec,
  title     = {{MVTec AD}---A Comprehensive Real-World Dataset for Unsupervised Anomaly Detection},
  author    = {Bergmann, Paul and Fauser, Michael and Sattlegger, David and Steger, Carsten},
  booktitle = {Proceedings of the IEEE/CVF Conference on Computer Vision and Pattern Recognition},
  pages     = {9592--9600},
  year      = {2019}
}

@inproceedings{roth2022patchcore,
  title     = {Towards Total Recall in Industrial Anomaly Detection},
  author    = {Roth, Karsten and Pemula, Latha and Zepeda, Joaquin and Sch{\"o}lkopf, Bernhard and Brox, Thomas and Gehler, Peter},
  booktitle = {Proceedings of the IEEE/CVF Conference on Computer Vision and Pattern Recognition},
  pages     = {14318--14328},
  year      = {2022}
}

@inproceedings{defard2020padim,
  title     = {{PaDiM}: A Patch Distribution Modeling Framework for Anomaly Detection and Localization},
  author    = {Defard, Thomas and Setkov, Aleksandr and Loesch, Ang{\'e}lique and Audigier, Romaric},
  booktitle = {Pattern Recognition. ICPR International Workshops and Challenges},
  pages     = {475--489},
  publisher = {Springer},
  doi       = {10.1007/978-3-030-68799-1_35},
  year      = {2020}
}

@misc{cohen2020spade,
  title        = {Sub-Image Anomaly Detection with Deep Pyramid Correspondences},
  author       = {Cohen, Niv and Hoshen, Yedid},
  year         = {2020},
  eprint       = {2005.02357},
  archivePrefix= {arXiv},
  primaryClass = {cs.CV}
}

@article{zhang2023aebad,
  title   = {Industrial Anomaly Detection with Domain Shift: A Real-World Dataset and Masked Multi-Scale Reconstruction},
  author  = {Zhang, Zilong and Zhao, Zhibin and Zhang, Xingwu and Sun, Chuang and Chen, Xuefeng},
  journal = {Computers in Industry},
  volume  = {151},
  pages   = {103990},
  doi     = {10.1016/j.compind.2023.103990},
  year    = {2023}
}

@article{candes2011rpca,
  title   = {Robust Principal Component Analysis?},
  author  = {Cand{\`e}s, Emmanuel J. and Li, Xiaodong and Ma, Yi and Wright, John},
  journal = {Journal of the ACM},
  volume  = {58},
  number  = {3},
  pages   = {11:1--11:37},
  doi     = {10.1145/1970392.1970395},
  year    = {2011}
}

@inproceedings{jeong2023winclip,
  title     = {{WinCLIP}: Zero-/Few-Shot Anomaly Classification and Segmentation},
  author    = {Jeong, Jongheon and Zou, Yang and Kim, Taewan and Zhang, Dongqing and Ravichandran, Avinash and Dabeer, Onkar},
  booktitle = {Proceedings of the IEEE/CVF Conference on Computer Vision and Pattern Recognition},
  pages     = {19606--19616},
  year      = {2023}
}

@inproceedings{cao2023adshift,
  title     = {Anomaly Detection Under Distribution Shift},
  author    = {Cao, Tri and Zhu, Jiawen and Pang, Guansong},
  booktitle = {Proceedings of the IEEE/CVF International Conference on Computer Vision},
  pages     = {6511--6523},
  year      = {2023}
}

@inproceedings{carvalho2023invariant,
  title     = {Invariant Anomaly Detection under Distribution Shifts: A Causal Perspective},
  author    = {Carvalho, Jo{\~a}o B. S. and Zhang, Mengtao and Geyer, Robin and Cotrini, Carlos and Buhmann, Joachim M.},
  booktitle = {Advances in Neural Information Processing Systems},
  volume    = {36},
  year      = {2023},
  url       = {https://proceedings.neurips.cc/paper_files/paper/2023/hash/b010241b9f1cdfc7d4c392db899cef86-Abstract-Conference.html}
}

@inproceedings{lendering2026subspacead,
  title     = {{SubspaceAD}: Training-Free Few-Shot Anomaly Detection via Subspace Modeling},
  author    = {Lendering, Camile and Akdag, Erkut and Bondarev, Egor},
  booktitle = {Proceedings of the IEEE/CVF Conference on Computer Vision and Pattern Recognition},
  pages     = {28557--28566},
  year      = {2026}
}

@inproceedings{gao2026residualfields,
  title     = {Anomaly-Related Residual Fields for Cross-domain Anomaly Detection},
  author    = {Gao, Kewei and Xie, Jiayi and Shen, Zhengda and Qin, Weijun and Jia, Lingxiang and Chen, Kejia and Feng, Zunlei and Bei, Yijun},
  booktitle = {Proceedings of the IEEE/CVF Conference on Computer Vision and Pattern Recognition},
  pages     = {35617--35627},
  year      = {2026}
}

@misc{han2026sparc,
  title        = {{SPARC}: Subspace Position-Aware Robust Few-Shot Calibration for Distribution-Shifted Industrial Anomaly Detection},
  author       = {Han, Seokhee and Chu, Seungjun and Nowak, Mateusz and Chin, Peter},
  year         = {2026},
  eprint       = {2608.18585},
  archivePrefix= {arXiv},
  primaryClass = {cs.CV}
}
